\documentclass{article}
\usepackage{spconf,amsmath,amssymb,graphicx}
\usepackage{ragged2e}
\usepackage[utf8]{inputenc}
\usepackage{multicol}
\usepackage{wrapfig}
\usepackage{mathtools}
\usepackage{fixltx2e,hyperref}
\usepackage[most]{tcolorbox}
\usepackage{booktabs, enumitem}
\usepackage{textpos}

\usepackage[pages=some,placement=top]{background}

\title{How to Fool Radiologists with Generative Adversarial Networks?\\ A Visual Turing Test for Lung Cancer Diagnosis}
\name{Maria J. M. Chuquicusma$^1$, Sarfaraz Hussein$^1$, Jeremy Burt$^2$, and Ulas Bagci$^1$}
\address{$^1$ Center for Research in Computer Vision (CRCV) at University of Central Florida, Orlando, FL.\\
$^2$ Radiology Department, Florida Hospital, Orlando, FL.}
\begin{document}
%
\backgroundsetup{contents=Accepted for publication in IEEE International Symposium on Biomedical Imaging (ISBI) 2018,color=black!100,scale=1.5,opacity=0.7,position={5.5,1.35}}
\BgThispage

\maketitle

\begin{abstract}
Discriminating lung nodules as malignant or benign is still an underlying challenge. To address this challenge, radiologists need computer aided diagnosis (CAD) systems which can assist in learning discriminative imaging features corresponding to malignant and benign nodules. However, learning highly discriminative imaging features is an open problem. In this paper, our aim is to learn the most discriminative features pertaining to lung nodules by using an adversarial learning methodology. Specifically, we propose to use unsupervised learning with Deep Convolutional-Generative Adversarial Networks (DC-GANs) to generate lung nodule samples realistically. We hypothesize that imaging features of lung nodules will be discriminative if it is hard to differentiate them (fake) from real (true) nodules. To test this hypothesis, we present Visual Turing tests to two radiologists in order to evaluate the quality of the generated (fake) nodules. Extensive comparisons are performed in discerning real, generated, benign, and malignant nodules. This experimental set up allows us to validate the overall quality of the generated nodules, which can then be used to (1) improve diagnostic decisions by mining highly discriminative imaging features, (2) train radiologists for educational purposes, and (3) generate realistic samples to train deep networks with big data.

\end{abstract}

\begin{keywords}
Lung nodules, Generated samples, Visual Turing Test, Computed Tomography (CT), Deep learning, Generative Adversarial Networks (GANs), Computer Aided Diagnosis (CAD) systems 
\end{keywords}

\vspace{-10 px}

\section{Introduction}
\label{sec:intro}
The primary cause of mortality today is cancer and annually it causes approximately 8.2 million deaths. In comparison to all cancers, lung cancer is the most lethal, as it causes roughly 1.59 million deaths per year~\cite{stewart2016world}.

Low dose computed tomography (CT) scans aid radiologists in screening lung cancer and can help early diagnosis as well. CT scans have the potential to save millions of lives if they are properly observed and interpreted in the early stages of lung cancer. One major challenge in automatic detection and classification of lung cancer is the variability in the appearance of lung nodules. Lung nodules are found in distinct shapes, sizes, intensities, locations, and other characteristics, regardless of being malignant or benign. Due to these characteristics, radiographic analysis of lung nodules can lead to an under or over diagnosis. Misinterpretation in this regard can lead to unnecessary surgeries and misdiagnoses. 
   
Computer-Aided Diagnosis (CAD) systems help in identifying discriminative features associated with malignant and benign lung nodules. Previous contributions by Hussein et al. have shown the effectiveness of supervised learning techniques for lung nodule classification. In their work, sagittal, axial, and coronal 2D patches of a CT scan are put into a 2D-Convolutional Neural Network (CNN)~\cite{hussein2017tumornet}. Extending this idea, Hussein et al. used supervised learning methodologies by fine-tuning a 3D-CNN on a visual dataset and by using multitask learning techniques to focus on specific attributes such as calcification, lobulation, sphericity, speculation, margin, and texture~\cite{hussein2017riskstratification}.
   

\begin{figure*}[t]
\centering
    \includegraphics[width=0.75\textwidth]{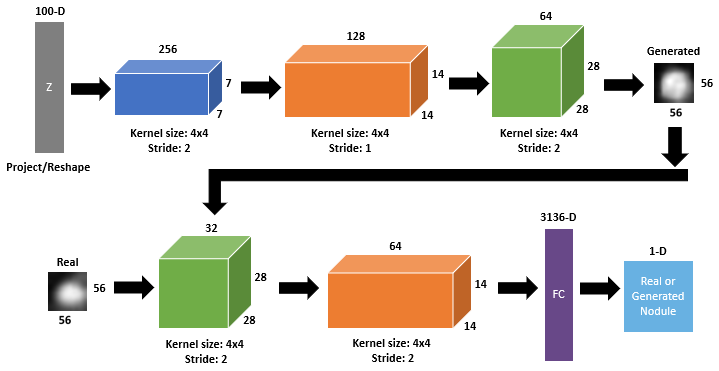}
	\caption{The DC-GAN architecture used for generating lung nodule samples. The generator consists of 3 convolutional layers and generates a 56 by 56 image sample. The discriminator, on the other hand, consists of 2 convolutional layers and takes a real and the generated image sample and outputs a probability score, which denotes whether the given images are real or generated.}
    \label{fig:DC-GAN}
\end{figure*}  

Recent works by Goodfellow et al. proposed Generative Adversarial Networks (GANs), a new framework that can learn discriminative features from images and generate realistic samples~\cite{IanGans}. Kamnitsas et al. further investigated this framework to learn domain invariant features for unsupervised automatic brain lesion segmentation~\cite{KonstaBrainLesions}. In a similar fashion, Schlegl et al. proposed AnoGANs to aid in the unsupervised detection of abnormalities in diseases found in imaging data. Their results showed that detecting anomalies, such as retinal fluid and hyper-reflective foci in brain tumors is now a possibility~\cite{SchleglUnsupervisedAnomaly}. These related works in the realm of both supervised and unsupervised learning, show that advanced machine learning methods, particularly deep learning and GANS, have made new enlightening discoveries in the medical imaging field. 

Specific to recent developments both in lung cancer screening and machine learning, \textbf{our contributions} in this study are the following: \textbf{(1)} We generate realistic lung nodules using GAN for the first time in the literature. Briefly, the presented GAN framework includes a generator which maps samples from an arbitrary latent distribution to data, and an adversarial discriminator that tries to distinguish between real and generated samples as accurately as possible. While the generator's goal is to "fool" the discriminator by producing samples which are as close to real data as possible, the discriminator's role is to learn the most powerful features in order not to be deceived.
\vspace{-1 px}
\textbf{(2)} We perform various Visual Turing tests with radiologists to qualitatively evaluate the generated samples and to determine the success of our approach.
\vspace{-1 px}
\textbf{(3)} Class-specific nodules (benign or malignant) are generated. We quantitatively measure the inter-observer variations, False Recognition Rate (FRR), and True Recognition Rate (TRR) to compute the goodness of the generated examples.
\vspace{-1 px}

We believe that generating high-quality lung nodules similar to the original (real) ones is of high importance for several reasons. Throughout our experiments, we generated samples for benign only, malignant only, and a mixture of both types. These generated images allow us to discern features specific to malignant and benign nodules, which could lead to better diagnosis, train radiology residents, and model complex deep learning classifiers. To verify that our samples are similar to the original samples, we conduct a Visual Turing test. Our objective is to convince radiologists that our generated samples are actually the original samples. This will give us qualitative and quantitative measures to evaluate our generated nodules. 

\section{Materials and Methods}
\label{sec:materials}
We use the publicly available lung nodule dataset, the Lung Image Database Consortium image collection (LIDC-IDRI)~\cite{armato2011lung}, for our experiments. This dataset contains 1018 lung cancer screening thoracic CT scans with annotations and malignancy ratings, where a rating of 1 denotes low malignancy and a rating of 5 represents high malignancy. Lung nodules that are of diameters greater than or equal to 3 mm were annotated by at most 4 radiologists. Nodules that were annotated by at least 3 radiologists are selected and nodules with a malignancy rating of 3 are excluded from selection to avoid high uncertainty. The final dataset used in this experiment consists of 1145 nodules; 635 are benign and 510 are malignant.

\begin{figure*}
\center
\includegraphics[width=.47\textwidth]{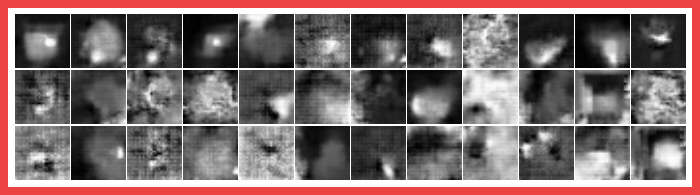}
\includegraphics[width=.47\textwidth]{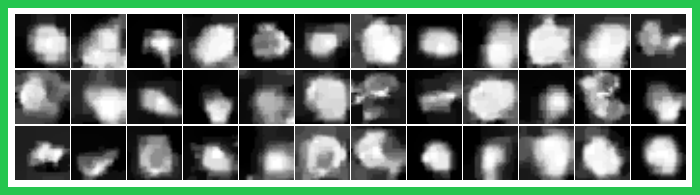}
 \caption{Generated low-quality (red) and high-quality (green) samples of lung nodules using a DC-GAN. In comparison with the low quality nodules, the high-quality nodules are more well-formed and defined.}
 \label{fig:bgsamples}
\end{figure*}

\begin{figure*}
\center
\includegraphics[width=0.47\textwidth]{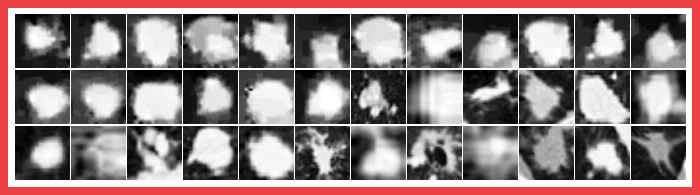}
\includegraphics[width=0.47\textwidth]{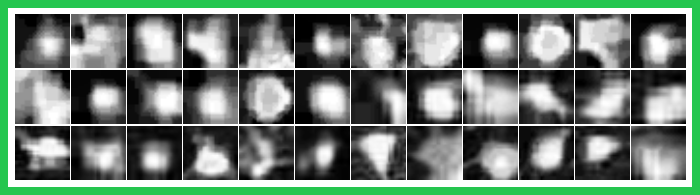}
 \caption{On the left (red), nodules 1-18 are generated by a DC-GAN and 19-36 are real malignant nodules. On the right (green), nodules 1-18 are generated by a DC-GAN and 19-36 are real benign nodules. Nodule numbering is from left to right.}
 \label{fig:classsamples}
\end{figure*}

\vspace{-5 px}
\subsection{Network Architecture}
Due to deep learning advances, technologies have the ability to meet almost the same level of image understanding as human beings. Specifically, GANs have been shown to produce exceptional results in this regard. To generate high-quality nodules, we propose to use Deep Convolutional Generative Adversarial Networks (DC-GANs). DC-GANs follow the same concept as the original GAN, which is composed of a generator $G$ and a discriminator $D$. Random noise is put into the generator and it generates a sample image. The objective is to eventually learn a distribution p\textsubscript{z} from this random noise. The goal for the generator is to generate samples, so as to deceive the discriminator as being original samples. The discriminator, then, classifies the original samples, which come from the p\textsubscript{data} distribution, and the generated samples, as fake or real. The discriminator and generator play a min-max game represented by the following value function:
\begin{equation}
\begin{split}
\min_{G} \max_{D} V(D,G) = \mathbb{E}_{x \sim p_{data}(x)} [log(D(x))] \\ +
\mathbb{E}_{z \sim p_{z}(z)} [log(1-D(G(z)))]
\end{split}
\end{equation}

DC-GANs augment the performance of the standard GAN by using convolutional layers along with batch normalization, removing fully connected layers, and utilizing ReLU, leaky ReLU, and tanh activation functions~\cite{radford2015unsupervised}. The architecture for the proposed experiments in this paper, shown in Figure~\ref{fig:DC-GAN} uses a generator comprised of 3 convolutional layers with batch normalization and a tanh activation function applied to the last layer. The discriminator includes 2 deconvolutional layers with leaky ReLU and a 3136-dimensional fully connected layer with a sigmoid activation function. 

We train the network to generate benign, malignant samples as well as a mixture of both malignant and benign samples. A batch size of 64 is used along with a learning rate of .0001 and .0002 for the discriminator and the generator. The network is trained for benign samples until 114,000 iterations, malignant samples until 110,000 iterations, and a mixture of malignant and benign samples until 99,000 iterations. There was no further improvement in the image quality after these numbers of iterations. Instead, the image quality started degrading, so we concluded those as our early stopping points for training.

\subsection{Visual Turing Test}
To compare the generated nodules with original nodules we propose 18 Visual Turing experiments. Our experiments were conducted with two radiologists, where radiologist 1 had 13 years of experience and radiologist 2 had 4 years of experience reading lung cancer scans. We conducted these set of experiments for benign, malignant, and a mixture of both benign and malignant nodules with both radiologists in radiology rooms to fully evaluate their perception of both real and generated nodules. Both radiologists were blinded of each other's evaluation of experiments and were not shown real or generated samples prior to experiments. Each Visual Turing experiment contains distinct 36 high-quality sampled generated nodules positioned in a 6 by 6 square grid; both radiologists were allowed to zoom or change the viewpoint of the image. Visuals did not contain patient information, background, or symptoms; only lung nodules from CT images were presented. CT images used were from the LIDC-IDRI dataset and were not prepared by the two radiologists who participated in our Visual Turing test. Anisotropic diffusion (Perona-Malik diffusion), which is a conventional noise removal filtering, was applied to images without distorting edge information while removing noise for better visualization~\cite{perona1990scale}. 
  
For each experiment of all generated nodules, all real nodules, and a mixture of generated nodules, we inform radiologists that the presented screen can contain anything (all real, all fake, or mixture). The proposed questions for radiologists were as follow:

\setlist[itemize]{leftmargin=*}
\begin{itemize}
\item[$\bullet$] \textbf{Experiments 1, 2, 3, 16, 17, 18 (All Nodules)}: If there is any, identify which nodule(s) are generated, which one(s) are real? 
\item[$\bullet$] \textbf{Experiments 4, 5, 6, 13, 14, 15 (Benign Nodules)}: If there is any, identify which nodule(s) are generated, which one(s) are real?  Furthermore, identify which nodule(s) are benign if there is any.
\item[$\bullet$] \textbf{Experiments 7, 8, 9, 10, 11, 12 (Malignant Nodules)}: If there is any, identify which nodule(s) are generated, which one(s) are real? Furthermore, identify which nodule(s) are malignant if there is any.
\end{itemize}

\begin{figure*}
\begin{center}

\includegraphics[scale=0.75]{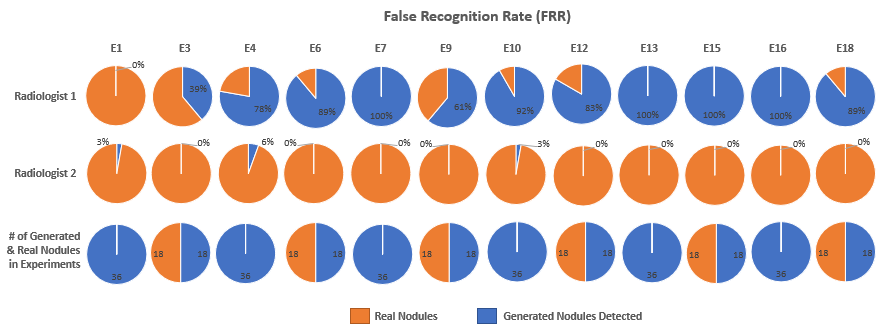}
\caption{The experiments above were used to calculate the FRR for the two radiologists and contain either all generated nodules or a mixture of real and generated nodules. The FRR shown in the pie charts indicate the percentage of nodules which radiologists recognized as generated (fake). E2, E5, E8, E11, E14, and E17 were excluded, since they contain all real nodules, hence they do not have results about fake or real lung nodule differences, but they do have results for malignant and benign lung nodules.}
  \label{FRR}
\end{center}
\end{figure*}

\begin{figure}
\centering
\includegraphics[width=0.33\textwidth]{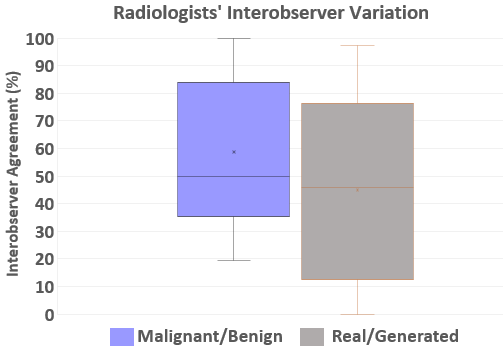}
 \caption{Radiologists' inter-observer variation for the respective cases of malignant/benign for the 36 samples within the experiments 4-9 and 13-18 and real/generated for the experiments 1-18.}
 \label{RIV}
\end{figure}

\vspace{-10 px}
\section{Results}
For our visual inspection and comparison we gather low-quality lung nodules along with high-quality samples as shown in Figure~\ref{fig:bgsamples}. Moreover, for our Visual Turing test, we also generated class-specific nodules and compared them with original nodules shown in Figure~\ref{fig:classsamples}. Lung nodule generated samples in both Figure ~\ref{fig:bgsamples} and ~\ref{fig:classsamples} all have anisotropic diffusion applied. 
To quantitatively measure our generated nodules, we calculate False Recognition Rate (FRR), shown in Figure~\ref{FRR}, and True Recognition Rate (TRR) of two radiologists' blinded evaluations of the experiments. FRR calculates the percentage of fake nodules recognized as fake by the radiologist whereas TRR calculates the percentage of real nodules recognized as real.

The FRR results show that we successfully convinced radiologist 1 that several of our generated nodules were real for 67\% of our experiments and radiologist 2 for 100\% of our experiments. Since radiologists face numerous challenges while classifying nodules, we also decided to investigate their TRR. Radiologist 1 had a mean TRR of 58.56\% while radiologist 2 had captured more true ones (93.52\%). To remain unbiased, we calculated radiologists' inter-observer variation, as summarized in Figure~\ref{RIV}, throughout experiments for malignant and benign nodules, as well as for real and generated nodules. Both radiologists had a mean inter-observer agreement for malignant and benign cases of 58.56\% and for real and generated cases of 44.91\%. These statistics indicate that the presented algorithm has the capability of generating high-quality lung nodules. These statistics also show that radiologists failed to distinguish between malignant and benign nodules accordingly. This performance can be ascribed to the successful generation of realistic images using the DC-GAN architecture.

\section{Discussion \& conclusion}
In this work, we generated realistic lung nodule samples using a DC-GAN architecture for the first time. The generated samples did manage to deceive radiologists in our experiments. The generated samples, however, do present limitations. One major limitation is that the features from both malignant and benign nodules could be present in a generated sample. This could ultimately cause difficulty in interpreting visualization for radiologists. Also, our generated samples would benefit from further improvement in terms of quality and variation. The results from our proposed experiments will help us readjust parameters and develop a more robust architecture for future experiments that can generate a larger number of high-quality realistic generated (fake) samples.

As future work, we also aim to expand our experimental setup by including more radiologists and experts. We will also explore image generation in the 3D space as CT scans are inherently volumetric. The use of a 3D network could bring in important volumetric information which would otherwise be lost in 2D image slices.

\bibliographystyle{IEEEbib}
\bibliography{refs}

\end{document}